\documentclass[conference]{IEEEtran}
\IEEEoverridecommandlockouts
\usepackage{cite}
\usepackage[hidelinks,breaklinks]{hyperref}
\usepackage{subfig}
\usepackage{amsmath,amssymb,amsfonts}
\usepackage{algorithmic}
\usepackage{graphicx}
\usepackage{textcomp}
\usepackage{xcolor}
\usepackage{multicol}
\usepackage{multirow}
\def\BibTeX{{\rm B\kern-.05em{\sc i\kern-.025em b}\kern-.08em
    T\kern-.1667em\lower.7ex\hbox{E}\kern-.125emX}}
\usepackage{comment}
\usepackage{xcolor}

\makeatletter
\newcommand{\linebreakand}{%
  \end{@IEEEauthorhalign}
  \hfill\mbox{}\par
  \mbox{}\hfill\begin{@IEEEauthorhalign}
}
\makeatother
  
\begin{document}

\title{Processing and Segmentation of Human Teeth from 2D Images using Weakly Supervised Learning}

\author{\IEEEauthorblockN{1\textsuperscript{st} Tomáš Kunzo}
\IEEEauthorblockA{\textit{Department of Applied Informatics} \\
\textit{Comenius University}\\
Bratislava, Slovakia \\
kunzo6@uniba.sk}
\and

\IEEEauthorblockN{2\textsuperscript{nd} Viktor Kocur}
\IEEEauthorblockA{\textit{Department of Applied Informatics} \\
\textit{Comenius University}\\
Bratislava, Slovakia \\
viktor.kocur@fmph.uniba.sk\\
0000-0001-8752-2685}
\and

\IEEEauthorblockN{3\textsuperscript{rd} Lukáš Gajdošech}
\IEEEauthorblockA{\textit{Department of Applied Informatics} \\
\textit{Comenius University}\\
Bratislava, Slovakia \\
lukas.gajdosech@fmph.uniba.sk\\
0000-0002-8646-2147}
\and

\linebreakand
\IEEEauthorblockN{4\textsuperscript{th} Martin Madaras}
\IEEEauthorblockA{\textit{Department of Applied Informatics} \\
\textit{Comenius University}\\
Bratislava, Slovakia \\
martin.madaras@fmph.uniba.sk\\
0000-0003-3917-4510}
}

\IEEEoverridecommandlockouts
\IEEEpubid{\makebox[\columnwidth]{979-8-3503-4353-3/23/\$31.00~\copyright2023 IEEE\hfill} \hspace{\columnsep}\makebox[\columnwidth]{ }}

\maketitle
\IEEEpubidadjcol
\begin{abstract}
Teeth segmentation is an essential task in dental image analysis for accurate diagnosis and treatment planning. While supervised deep learning methods can be utilized for teeth segmentation, they often require extensive manual annotation of segmentation masks, which is time-consuming and costly. In this research, we propose a weakly supervised approach for teeth segmentation that reduces the need for manual annotation. Our method utilizes the output heatmaps and intermediate feature maps from a keypoint detection network to guide the segmentation process. We introduce the TriDental dataset, consisting of 3000 oral cavity images annotated with teeth keypoints, to train a teeth keypoint detection network. We combine feature maps from different layers of the keypoint detection network, enabling accurate teeth segmentation without explicit segmentation annotations. The detected keypoints are also used for further refinement of the segmentation masks. Experimental results on the TriDental dataset demonstrate the superiority of our approach in terms of accuracy and robustness compared to state-of-the-art segmentation methods. Our method offers a cost-effective and efficient solution for teeth segmentation in real-world dental applications, eliminating the need for extensive manual annotation efforts.

\end{abstract}


\section{Introduction}

Teeth segmentation is a critical task in dental image analysis, enabling accurate diagnosis, treatment planning, and monitoring of dental conditions. Deep learning approaches have shown promise in teeth segmentation, but they often rely on extensive manual annotations, which can be time-consuming and costly to obtain. Therefore, there is a need for automated teeth segmentation methods that can effectively segment teeth from RGB images without the need for explicit segmentation mask annotations.

In this paper, we propose a weakly supervised approach for teeth segmentation that aims to alleviate the requirements of manual annotation. Our method leverages the output heatmaps and intermediate feature maps from a keypoint detection network to guide the segmentation process. To train the keypoint detection network we have annotated a new TriDental dataset of 3000 images of the oral cavity with teeth keypoints. The trained keypoint detection network learns to discern significant features of teeth. Therefore, we propose to utilize the feature maps from the backbone of the keypoint detection network to guide segmentation.

To perform segmentation, we combine feature maps from different layers of the keypoint detection network to capture both multi-scale information about the teeth. This multi-scale feature fusion allows us to generate accurate masks of teeth without the need for explicit segmentation annotations. We also use the detected keypoints for further refinement of the obtained masks. Our approach is generic and can adapt to different dental image datasets, making it versatile and applicable in various dental image analysis scenarios.

To evaluate the effectiveness of our proposed approach, we conduct experiments on the TriDental dataset. We compare our results with the state-of-the-art segmentation methods and demonstrate the superiority of our approach in terms of accuracy and robustness. Our method not only eliminates the need for extensive manual annotation efforts but also offers a cost-effective and efficient solution for teeth segmentation in real-world dental applications.

\section{Related Work}

Standard deep learning segmentation methods \cite{MaskRCNN, unet} require large amounts of images annotated with segmentation masks. This type of annotation requires a long time to produce thus making the cost of generating such datasets prohibitively expensive. In this section, we review and analyze the existing literature on image segmentation methods that do not require such extensive mask annotations. 

\subsection{Weakly supervised semantic segmentation}

One way to circumvent the need to obtain fully labeled datasets is to use rely on weakly supervised methods. These methods rely on training data which does not contain the full segmentation masks, but only other type of annotations which are easier to produce such as class labels or object keypoints.

Classification networks trained with image-level labels can produce class response maps (or class activation maps), which correspond to the classification confidence of different image regions. Moreover, the local maxima, or peaks, of these regions often resemble highly contributing parts of different instances of objects in the image. To obtain these peaks authors of \cite{peak-response} first converted the classifier network into a fully convolutional network, which preserves spatial information and produces class response maps in a forward pass. Then they constructed a peak stimulation layer that is inserted after the final layer, which outputs class-wise confidence scores, meaning that each peak is associated with a class score. Because the class response maps are usually of small size and low resolution, they alone are insufficient for segmentation. That is why the authors propose to further backpropagate the stimulated class-wise confidence scores (peaks) to generate more informative regions of each object, referred to as peak response maps. The backpropagation of peaks is a probabilistic process. The top-down relevance of each object is formulated at the bottom layer as the object's probability of being visited in a random walk. With this probability propagation, the most relevant spatial locations of each object can be localized and used to create the peak response maps.  These maps are fine-detailed and can be further exploited for instance segmentation of objects.

Authors of \cite{ThePoint} proposed a supervision regime for semantic segmentation using human annotations in the form of keypoints indicating object locations. By incorporating point supervision into the training loss of a segmentation CNN framework, they managed to significantly improve semantic segmentation accuracy. To address the challenge of inferring the full object extent from points, they introduce an objectness prior that guides the training process by providing probabilities for pixels belonging to objects. This approach achieved a significant improvement compared to image-level class-based supervision.

\subsection{Self-supervised segmentation}

A different approach requires no training annotations, instead relying on self-supervision. CutLER (cut-and-learn) \cite{cutler} relies only on images for training and can be directly employed to perform complex segmentation and detection tasks over a wide range of domains. CutLER starts with standard self-supervised features~\cite{DINO} which are used by a mechanism called MaskCut to obtain multiple coarse segmentation masks. A simple loss-dropping strategy, which is robust to objects missed by MaskCut, is then used to train detectors using the coarse masks. Despite learning from these coarse masks, the detectors ‘clean’ the ground truth and produce masks that are better than the coarse masks used to train them. Therefore, multiple rounds of self-training on the models’ own predictions allow it to evolve from capturing the similarity of local pixels to capturing the global geometry of the object, thus producing finer segmentation masks. Since this segmentation mask has been trained on the unlabeled images from the large ImageNet dataset~\cite{imagenet} it is capable of segmenting a large variety of objects.

\subsection{Prompt based segmentation}

A different approach to the problem of lack of annotated masks is to use models capable of segmenting images of any category without training on category-specific data based solely on prompts. Such prompts should specify what part of the image should be segmented and the method should output a valid segmentation mask. The prompts could be keypoints, bounding boxes, partial masks, or natural language descriptions of the objects. The Segment Anything model \cite{SAM} is a recent advancement in this field. The model consists of a prompt encoder, an image encoder, and a lightweight mask decoder. The separation of the model into image encoder and prompt encoder is a key principle because this way one image can be used with different prompts while preserving the same image embedding and therefore saving computational time. Motivated by scalability and powerful pre-training methods the authors use a visual transformer altered to process high-resolution images \cite{sam_vit} as the image encoder. On the output end, the mask decoder is also built from transformer \cite{transformer} architecture. It is a modified transformer decoder block with a mask prediction head. This model was trained on a large dataset containing over 11 million images annotated with 1 billion segmentation masks.

\section{Dataset}

\begin{figure}

\centering
\begin{tabular}{c c c}
\includegraphics[height=1.55cm]{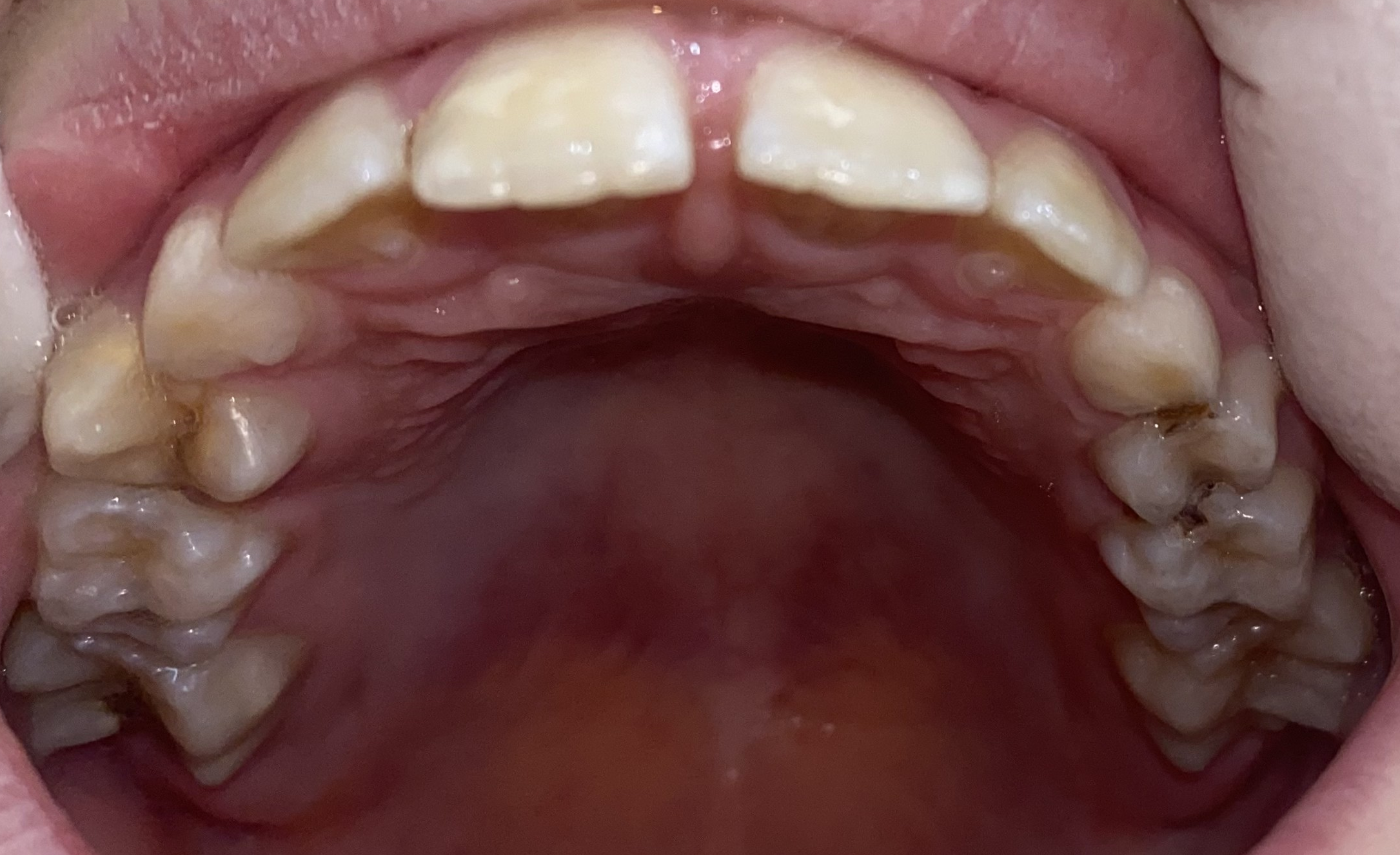} &
\includegraphics[height=1.55cm]{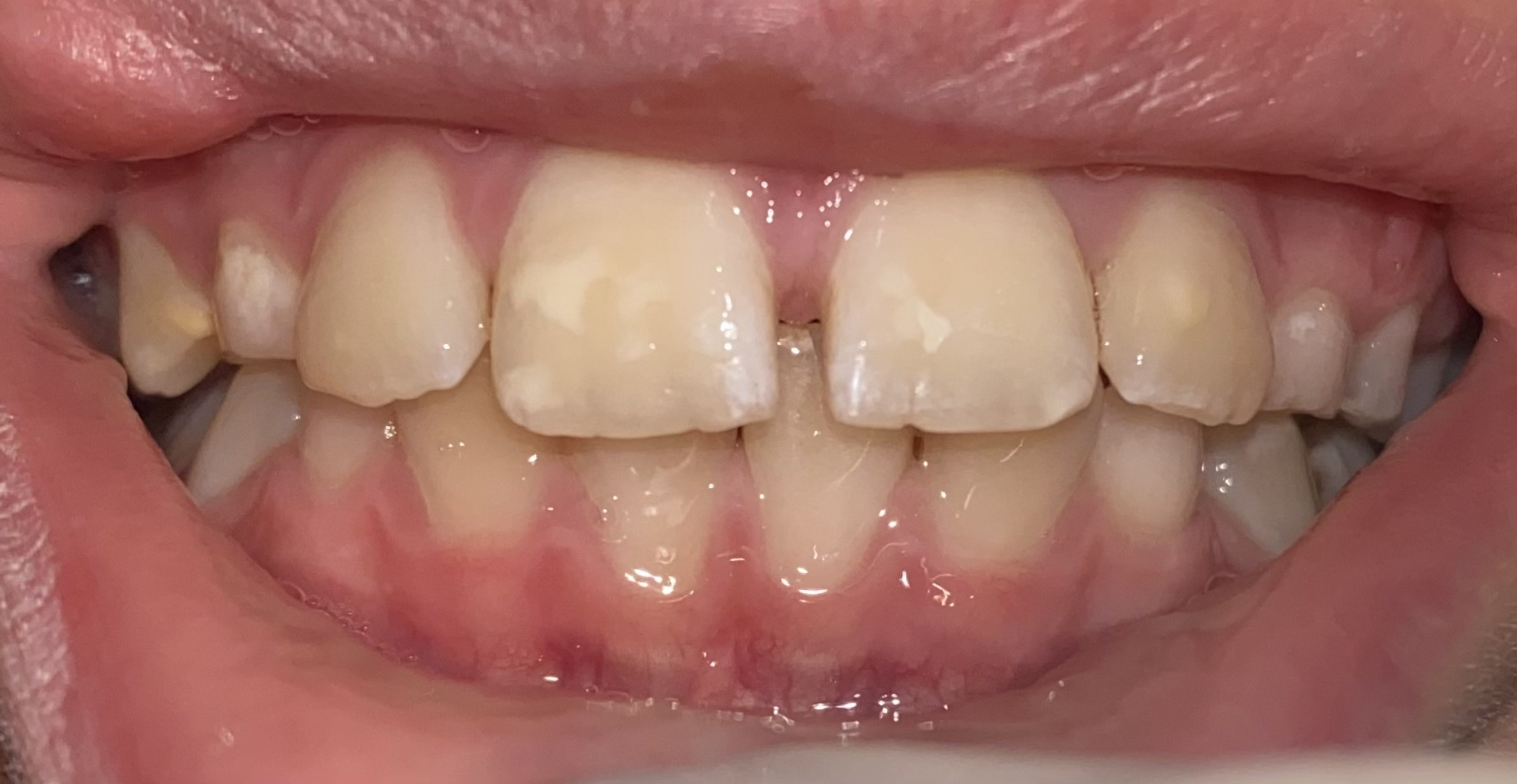} &
\includegraphics[height=1.55cm]{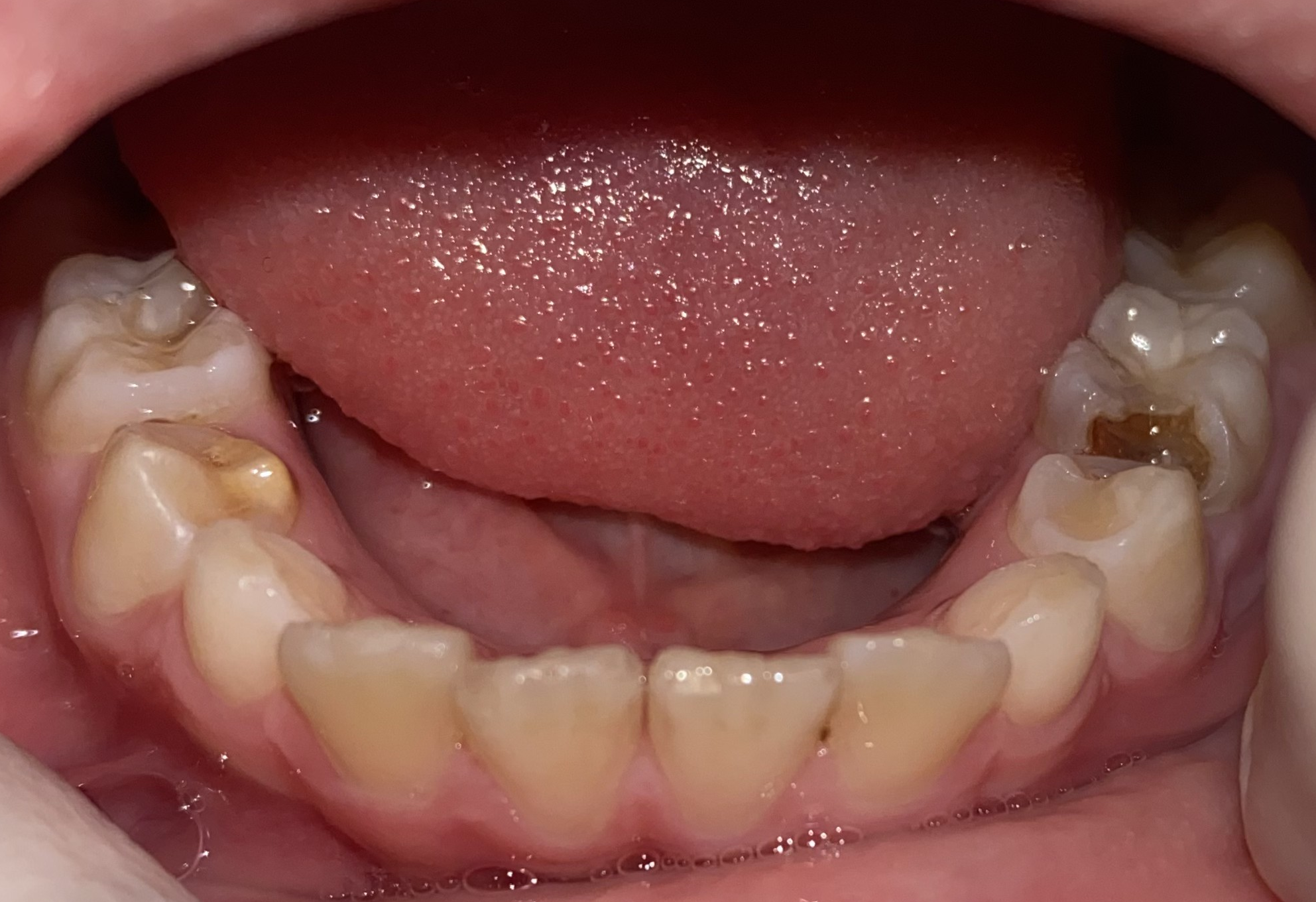} \\
Upper &Front & Lower
\end{tabular}
\caption{Example of three different views of the same oral cavity representing a single sequence in the TriDental dataset.
\label{fig:dataset_views}}
\end{figure}

To enable research of computer vision methods in dentistry we have collected and annotated the TriDental dataset consisting of RGB images of oral cavities. The images were collected by medical professionals. The dataset contains images of oral cavities of 1000 individuals. Each individual's oral cavity was photographed from three distinct views as shown in Fig. \ref{fig:dataset_views} resulting in 3000 RGB images.

To annotate the data we have created an online annotation tool that could be used by annotators. Each sequence of three views could be loaded at once. The annotators then provide keypoints for each tooth in each view. Since keypoint location on a tooth is not well-defined the annotators were instructed to choose keypoint locations corresponding to the centers of teeth. Annotated keypoints can be seen in Fig. \ref{fig:keypoint-results}.

For 10 samples from the dataset, we have also manually created masks of teeth resulting in 30 segmentation masks. We use this subset of the TriDental dataset to evaluate the segmentation approaches in section \ref{sec:segmentation_results}.

\section{Keypoint Detection}

\label{sec:keypoint}

In this section we will describe the proposed method for teeth keypoint detection. Our method is based on the CenterNet object detection network~\cite{Centernet}. We use the output keypoints and the learned features from the trained keypoint detection network to perform teeth segmentation. This approach is more closely described in section \ref{sec:proposed}. We also use the keypoints estimated by the network to provide input for the Segment Anything segmentation method \cite{SAM} which we describe in section \ref{sec:sam}.

\subsection{Keypoint detection network}

We base our keypoint estimation network on the CenterNet object detector~\cite{Centernet}. CenterNet detects objects by estimating the positions of their centers along with additional regression of objects widths, heights and center offsets. The centers are estimated in a form of a heatmap with Gaussian peaks representing the individual object centers. The positions of the centers are refined using the output of the offset regression maps. To detect only the keypoints we remove the output heads responsible for width and height estimation and keep only the keypoint heatmap and offset regression maps on the output. We used the ResNet18~\cite{resnet} network as the backbone. The default model outputs a $128 \times 128$ heatmap. We also performed an experiment by adding two more transpose convolution layers at the end of the model to obtain an output heatmap with the size of $512 \times 512$.

To train the network we used the TriDental dataset. We split the data into a 80/10/10 training/val/test split. Before feeding the images into the network we resized them to $512$ px $\times$ $512$ px resolution. We used batch size of 32 and trained the models for 40 epochs using the standard training schedule proposed by the authors in \cite{Centernet}.

Since the upper and lower view images are significantly different from the frontal views images we have performed multiple experiments by training the network on all views simultaneously, each view individually and only on the upper and lower view images.

\subsection{Results}

\begin{figure*}[ht!]
    \includegraphics[width=0.19\textwidth]{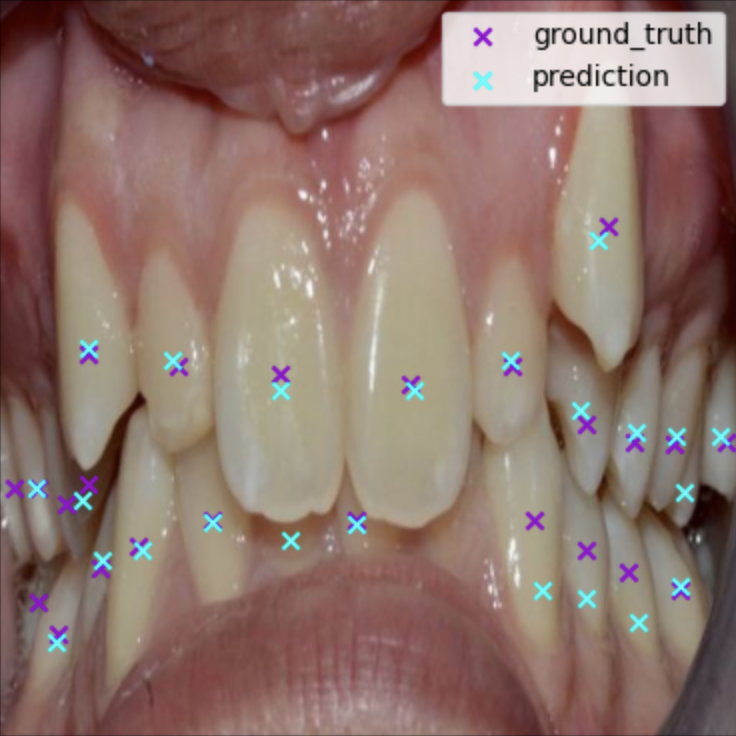} \hfill
    \includegraphics[width=0.19\textwidth]{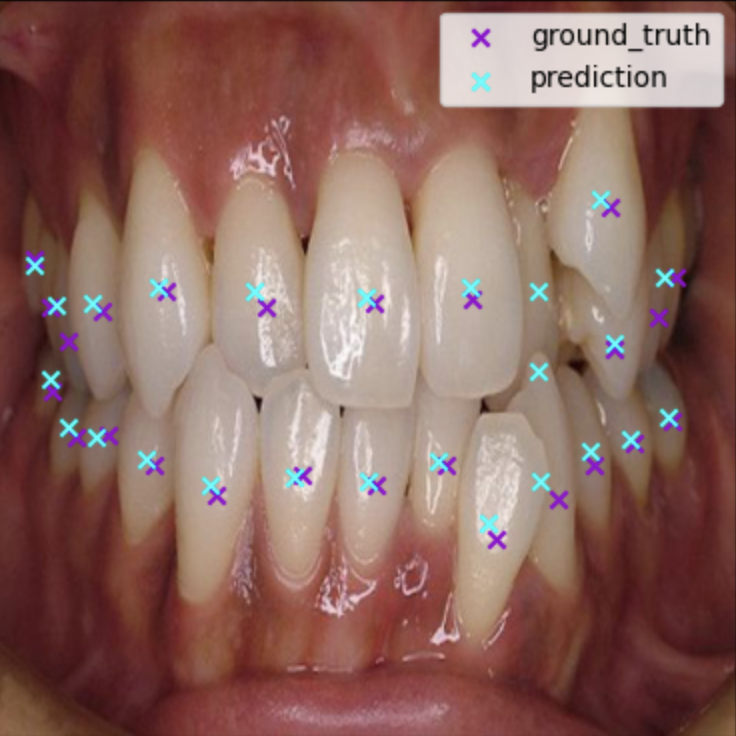} \hfill
    \includegraphics[width=0.19\textwidth]{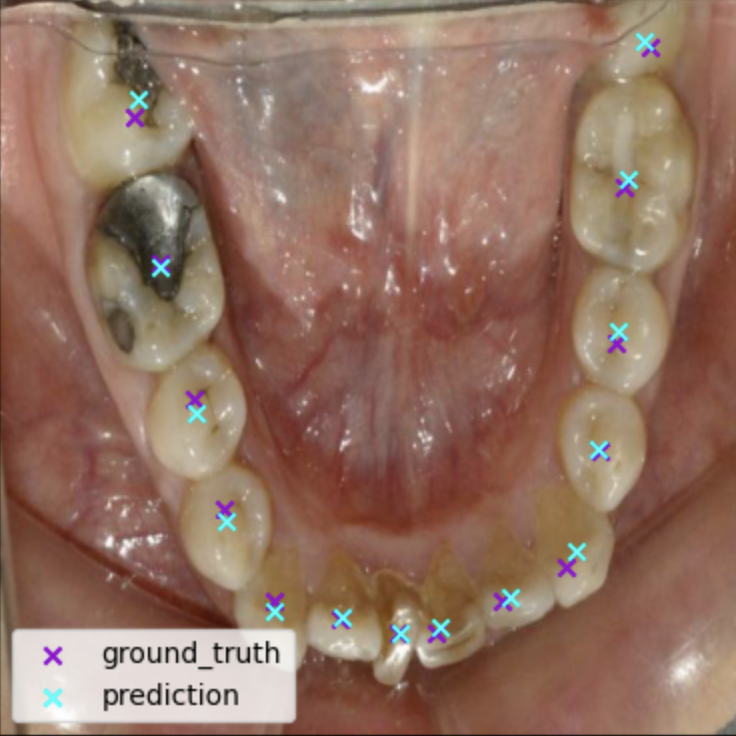} \hfill
    \includegraphics[width=0.19\textwidth]{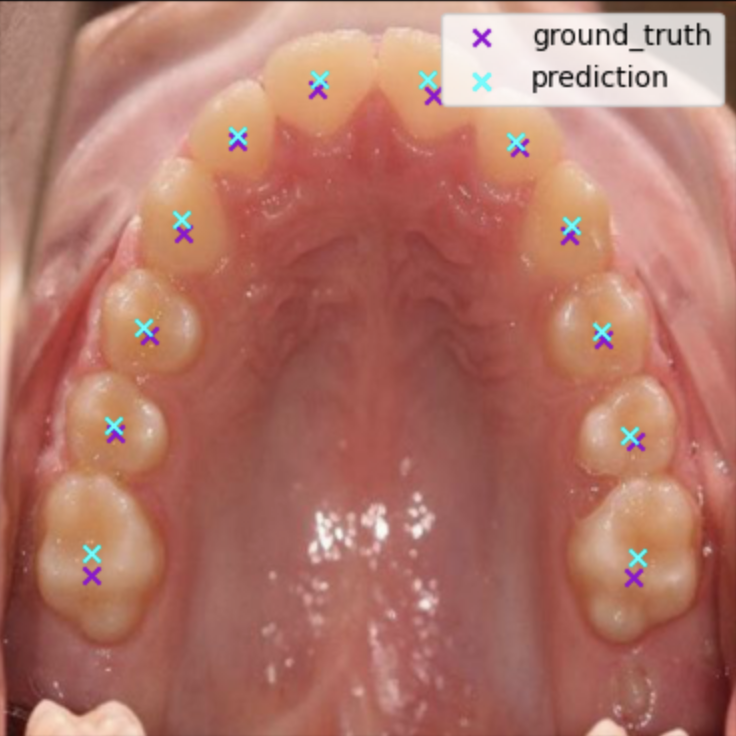} \hfill
    \includegraphics[width=0.19\textwidth]{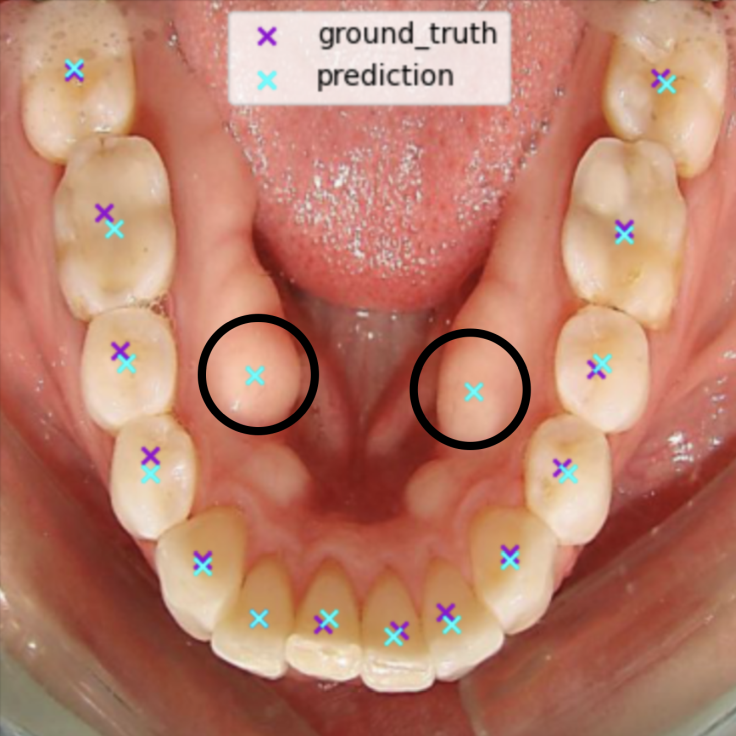} \hfill
    
    \caption{Example of our keypoint prediction with network with $512 \times 512$ output heatmap trained on all views. Dark purple markers represent ground-truth keypoints, turquoise markers represent predictions. The image on the right shows two false positive detections.}
    \label{fig:keypoint-results}%
\end{figure*}

\begin{table*}[ht!]
\centering
\caption{The mean and absolute distances of estimated keypoints to ground truth keypoints on the test set of TriDental. Estimated keypoints were associated with ground truth annotations using the Hungarian algorithm~\cite{hungarian}. The columns indicate which views the model was trained and evaluated on. The reported distances are measured in the original image coordinates with size $512$ px $\times 512$ px.}
\begin{tabular}{ |c|c|c|c|c|c|c| } 
 \hline
  Output heatmap size & Metric & All images & Lower only & Front only & Upper only & Lower + Upper  \\
 \hline
\multirow{ 2}{*}{128x128} 
& mean (px) & 9.07 & 14.86 & 16.17 & \textbf{6.82} & 8.83 \\ 
& median (px) & 6.43 & 8.5 & 12.31 & 6.45 & 7.93 \\ 
\hline
\multirow{ 2}{*}{512x512}
& mean (px) & 13.92 & \textbf{7.39} & \textbf{13.65} & 6.97 & \textbf{7.29} \\ 
& median (px) & 6.7 & \textbf{5.68} & \textbf{5.7} & \textbf{6.14} & \textbf{5.52} \\ 
 \hline
\end{tabular}
\label{table:keypoint-results-basic}
\end{table*}

The ground truth keypoints and keypoints keypoints estimated by the model trained on all three views with the $512 \times 512$ output heatmap are shown in Fig. \ref{fig:keypoint-results}. To gauge the accuracy of the estimated keypoints we match them to the ground truth keypoints using the Hungarian algorithm \cite{hungarian}. We report the mean and median distances of the matched pairs. The results for all of the evaluated models are shown in Table~\ref{table:keypoint-results-basic}.

The mean distance results could be skewed by a few bad examples. Therefore, we deem the median values to be generally more indicative of the model performance. The results for the model with the larger output size indicate that using a separate model for the front view and another for both the upper and lower view is desirable.

\begin{figure*}[ht!]
    \centering
    \subfloat[\centering Results from model with output size 128x128 pixels.]{{\includegraphics[width=0.47\textwidth]{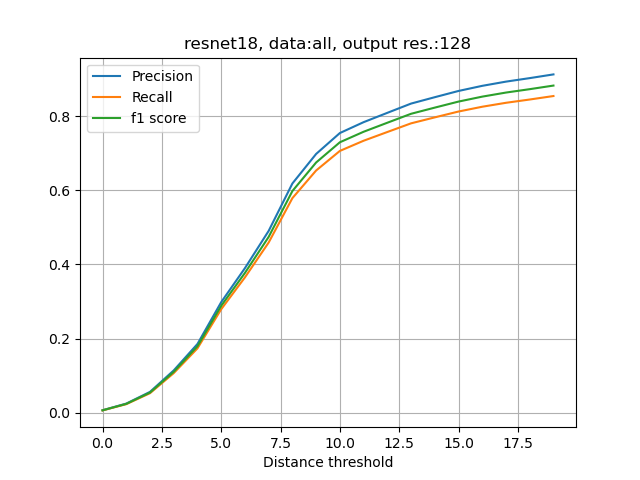} }\label{resnet-all-128}}%
    \hfill
    \subfloat[\centering Results from model with output size 512x512 pixels.]{{\includegraphics[width=0.47\textwidth]{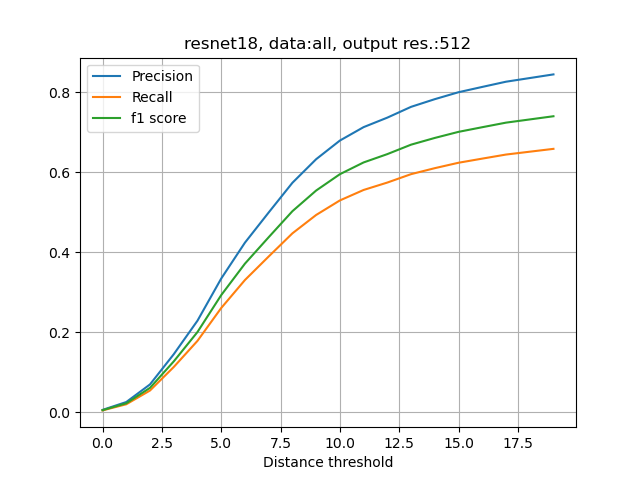} }\label{resnet-all-512}}%
    \caption{Precision, recall, and f1 score are displayed on the y-axis as the distance threshold is varied on the x-axis. We show the results on the test set of TriDental for the models trained on all images. 
    }
    \label{fig:resnet-all-0}%
\end{figure*}

Since calculating the mean and median distances is only possible for the keypoints which have been successfully paired by the Hungarian algorithm, relying only on these metrics may be misleading. Additionally, the positions of the ground truth keypoints for individual teeth are not defined precisely, therefore small distances between the ground truth and estimated keypoitns can be considered as insignificant in terms of accuracy when the system is used in practice. We therefore also calculated the precision, recall and f1 score for each model. The motivation behind these three metrics is to include the fact that the model can sometimes predict different number of keypoints than the ground-truth contains. To define a true positive match of an estimated keypoint paired with a ground truth one we use a distance threshold. When the distance of an estimated keypoint to a matched ground truth keypoint is lower than a given threshold we consider the keypoint to be true positive. Otherwise, we consider the estimated keypoint as a false positive and the ground truth keypoint as a false negative. When an estimated keypoint is not matched to any ground truth keypoint we consider it to be a false positive. As a false negative we consider the opposite case. We visualize the metrics by varying the threshold as shown in Fig. \ref{fig:resnet-all-0}.

The results show that our approach to teeth keypoint detection provides accurate results. We leverage the capabilities of the trained keypoint detection model in order to guide segmentation of teeth without requiring any mask ground truth annotations. Our weakly supervised teeth segmentation approach is described in the next section.

\section{Proposed Method for Weakly Supervised Teeth Segmentation}

\label{sec:proposed}

In this section we describe our proposed approach for weakly supervised teeth segmentation. The goal of this task is to obtain a method capable of generating masks of teeth from RGB images of the oral cavity. Our approach utilizes the output heatmaps and the feature maps from the keypoint detection network described in section \ref{sec:keypoint} to guide segmentation. No further training is performed as our method does not require any segmentation mask annotations.

\subsection{Feature maps}


\begin{figure}[t!]
    \centering
    \includegraphics[width=\linewidth]{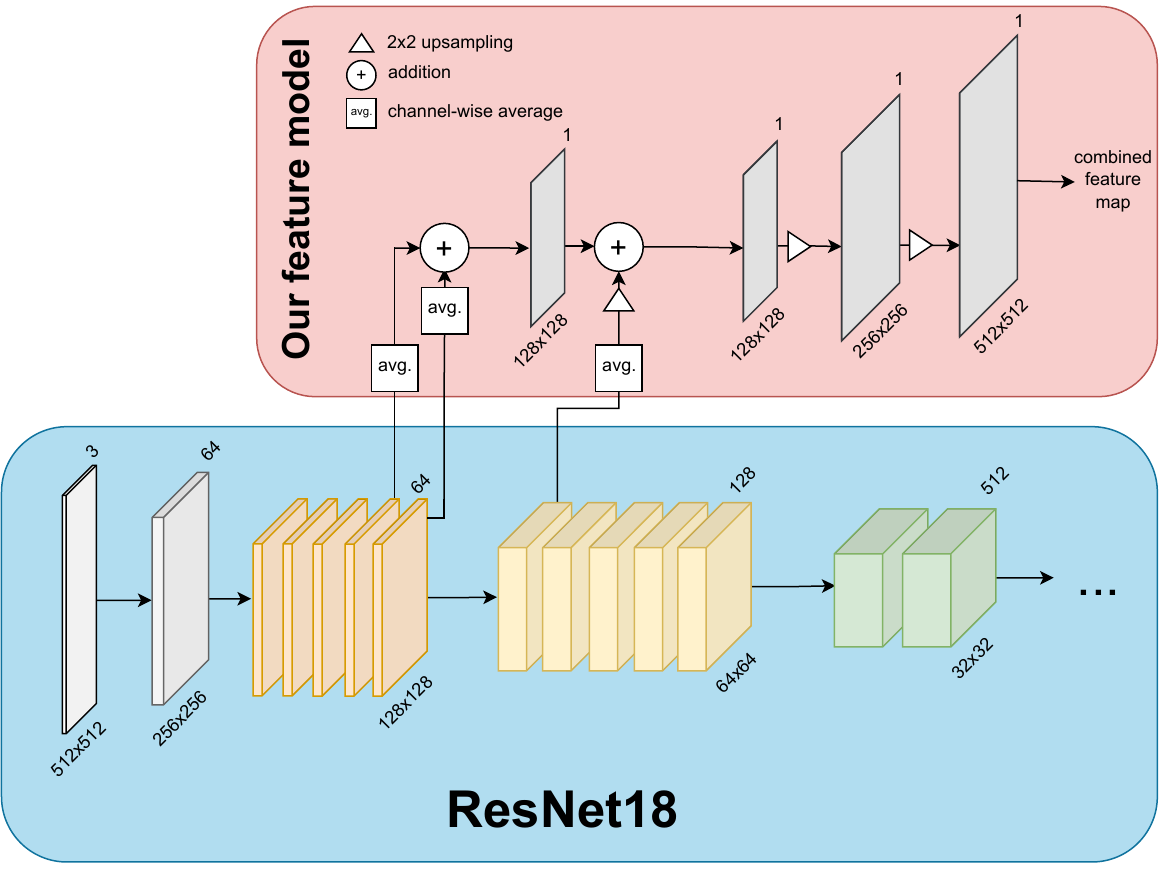}
    \caption{We combine several feature maps from the ResNet18~\cite{resnet} backbone of the trained keypoint detection network into a single combined feature map which we further process to obtain segmentation masks. The obtain the combined feature map, feature maps from three different layers of the model are averaged, added together, and upsampled to the resolution of the input image.}
    \label{fig:feature-model}
\end{figure}

The keypoint detection network learns various features that enable it to perform its tasks. These learned features are contained in the weights (filters) of the trained model. When an image is fed into the keypoint detection network the filters are applied to calculate the feature maps at each layer of the network. These feature maps carry information ranging from low-level features such as edges to higher-level semantic features.


Inspired by Feature Pyramid networks \cite{fpn} and IRNet architectures \cite{IRNet}, we combine feature maps from three different layers of the backbone of the keypoint detection model and get a rich feature representation of teeth in the image. We use the model with $512 \times 512$ output heatmap trained on all views. The model's Resnet18~\cite{resnet} backbone has three different sections. We consider the last two convolutional layers of the first section and the first layer of the second section. We combine the feature maps by calculating their channel-wise average as we noticed that in general, the feature maps contain positive activations in the areas where the teeth are located. The feature maps from the second section have smaller dimensions compared to the feature maps in the first section so before combining the maps we upsample the feature map from the second section. After combining the maps we upsample the resulting feature map to $512 \times 512$ resolution. Fig. \ref{fig:feature-model} shows how the feature maps from the backbone of the keypoint detection network are processed to obtain the final combined feature map.

\subsection{Postprocessing}

\begin{figure*}[ht!]
    \centering
    \subfloat[\centering Combined feature map]{{\includegraphics[width=0.22\textwidth]{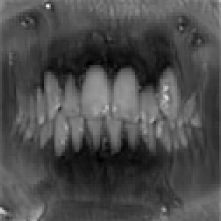} }\label{fig:our-fm}}%
    \quad
    \subfloat[\centering Thresholded]{{\includegraphics[width=0.22\textwidth]{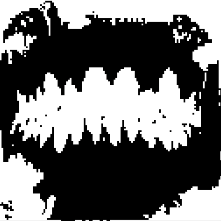} }\label{fig:our-thresh}}%
    \quad
    \subfloat[\centering Morph. closed]{{\includegraphics[width=0.22\textwidth]{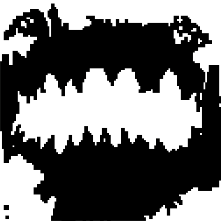} }\label{fig:our-closed}}%
    \quad
    \subfloat[\centering Refined with CRF]{{\includegraphics[width=0.22\textwidth]{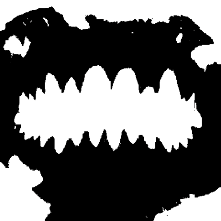} }\label{fig:our-crf}}%
    
    \subfloat[\centering Watershed distances.]{{\includegraphics[width=0.22\textwidth]{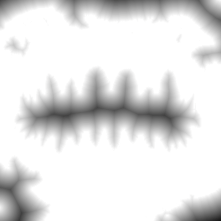} }\label{fig:our-distances}}%
    \quad
    \subfloat[\centering Predicted keypoints area]{{\includegraphics[width=0.22\textwidth]{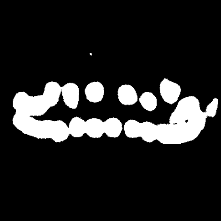} }\label{fig:our-keypoints-area}}%
    \quad
    \subfloat[\centering Distances with keypoints area]{{\includegraphics[width=0.22\textwidth]{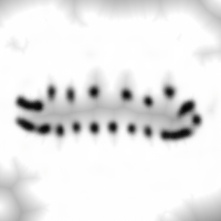} }\label{fig:our-distances-keypoints-area}}%
    \quad
    \subfloat[\centering Output mask]{{\includegraphics[width=0.22\textwidth]{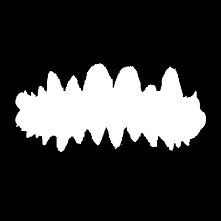} }\label{fig:our-final-mask}}%
    \caption{Post-processing pipeline of our method. The combined feature map is postprocessed using Otsu's thresholding~\cite{otsu}, morphological operations, CRF~\cite{CRF} and a modified version of the watershed algorithm using the output heatmaps of our keypoint detection method to obtain the final segmentation masks. }
    \label{fig:our-method-pipeline}%
\end{figure*}

The resulting feature map has to further be processed to produce the output mask. First, we pass our generated feature map into the Otsu’s automatic thresholding algorithm \cite{otsu} and then we use morphological closing operation to artifacts (this can be seen in \mbox{Fig.~\ref{fig:our-fm}-\ref{fig:our-closed}}). 

Since our combined feature map was created using upsampling layers multiple times, its resolution is quite low. Hence, to increase the resolution of our segmentation masks we use the CRF algorithm \cite{CRF} without any fine-tuning. This algorithm enhances and smooths the rough edges of our segmentation mask, see Fig.~\ref{fig:our-crf}. 

However, there can still be multiple segmentation artifacts in this mask, so we make use of watershed algorithm \cite{watershed} combined with predicted keypoints from the keypoint detection model. The watershed transformation has several steps. It starts by converting the input into a topological space with peaks and valleys. In our case, since the input is already a binary image, it converts the white regions to peaks by calculating the distance of each white pixel to the nearest black pixel. This way, the center areas of each tooth will have the highest values, as they are further away from black pixels than other pixels, see Fig.~\ref{fig:our-distances}. Having constructed such peaks, the algorithm then thresholds the space to obtain only peaks of a certain height. Then it labels each peak and its surrounding area with the value of that peak. At last, it inverts the space, so the peaks become basins and flood it. 

We modified two steps of this algorithm. Firstly, as the topological representation gets computed, we add to it the heatmaps predicted by our keypoint detection model to make the peaks significantly higher, see Fig.~\ref{fig:our-keypoints-area} and Fig.~\ref{fig:our-distances-keypoints-area}. The reason for this is that the segmentation artifacts may also become as high as the teeth regions. By applying our predicted heatmaps, the height difference between teeth areas and everything else becomes larger, so it is much easier for peak thresholding to select only peaks corresponding to teeth. However, in certain cases, where there are large white artifacts, they can become almost as high as the teeth regions. Therefore, we also use the number of predicted keypoints from the heatmap as an additional guide for the algorithm to not only select peaks above a certain threshold but also a certain number of them. The watershed algorithm then outputs the final segmentation mask. The whole pipeline is depicted in Fig.~\ref{fig:our-method-pipeline}.

\section{Teeth Segmentation Evaluation}

In this section we evaluate our weakly-supervised segmentation approach on the subset of TriDental dataset with segmentation labels. We first describe a preprocessing step used in some experiments. Then we briefly describe other segmentation methods we used to compare the results of our method. Finally, we discuss the evaluation results.

\subsection{Inpainting}

Images in the TriDental dataset contain significant reflections. We have thus decided to consider a prerprocessing step that removes reflections from the data. To remove reflections we use the Navier-Stokes inpainting algorithm~\cite{navier}. This algorithm inpaints the bright white spots in our images with the surrounding colors.

\subsection{Baselines}

To gauge the accuracy of our method we compare it to several baselines. As comparisons, we used two simple analytical methods and two deep-learning-based methods.

The first method relies on Otsu's thresholding \cite{otsu}. To apply the method the original image gets converted to grayscale color space, then a rough binary mask is created using Otsu's thresholding \cite{otsu}. These rough masks is then further refined with morphological closing and hole-filling algorithms to get rid of artifacts. 

The second baseline analytical approach is based on HSV thresholding. To apply the method we first convert the RGB images into the HSV representation. Then we use manually selected lower and upper thresholds for each of the three values to obtain the final segmentation mask. To obtain the best results we created two separate masks. The first mask selected the teeth boundary and the second mask selected the rough region of teeth. We merged these two masks to obtain the final segmentation mask.

\label{sec:sam}

We also evaluate the pre-trained Segment Anything model \cite{SAM}. We use the Segment Anything model in two different modes. As the first mode we tried the mode which does not receive any specific prompt and tries to automatically segment the foreground. In the second regime, we used the keypoints obtained from our keypoint detection network with $512 \times 512$ output heatmap trained on all views as prompts. 

Additionally, we also report the results for the CutLER model~\cite{cutler} with the DINO~\cite{DINO} backbone.

\subsection{Results}
\label{sec:segmentation_results}

\begin{table}
\centering
\caption{The intersection-over-union of estimated masks compared to the ground truth mask annotations on the subset of the TriDental dataset with annotated segmentation masks for the selected methods. The values are reported individually for each view. Bold values indicate the best-achieved results. The method SA indicates the use of the Segment Anything model~\cite{SAM} without a specific prompt. The method SA + KP indicates the use of the Segment Anything model~\cite{SAM} with the keypoints from our keypoint detection method.
\label{tab:segmentation_results}}
\begin{tabular}{|c|c|c|c|c|c|}
\hline
Method                                  & Inpainting & Lower & Front & Upper \\ \hline
\multirow{2}{*}{Otsu \cite{otsu}}       &               & 0.501 & 0.085 & 0.382 \\ 
                                        & \checkmark    & 0.496 & 0.278 & 0.377 \\ \hline
\multirow{2}{*}{HSV thresholding}       &               & 0.601 & 0.581 & 0.441 \\ 
                                        & \checkmark    & 0.601 & 0.553 & 0.443 \\ \hline
\multirow{2}{*}{CutLER \cite{cutler}}   &               & 0.413 & 0.307 & 0.269 \\ 
                                        & \checkmark    & 0.320 & 0.316 & 0.240 \\ \hline
\multirow{2}{*}{SA-auto \cite{SAM}}     &               & 0.369 & 0.377 & 0.367 \\ 
                                        & \checkmark    & 0.317 & 0.311 & 0.358 \\ \hline
\multirow{2}{*}{SA\cite{SAM} + KP (ours)}    &               & \textbf{0.876} & 0.555 & \textbf{0.909} \\ 
                                        & \checkmark    & 0.830 & 0.477 & 0.799 \\ \hline
\multirow{2}{*}{Ours}                   &               & \textbf{0.876} & \textbf{0.775} & 0.669 \\ 
                                        & \checkmark    & 0.846 & 0.717 & 0.740 \\ \hline
\end{tabular}
\end{table}

\begin{figure}[htbp]
    \centering     
    \subfloat[\centering Prediction on lower jaw.]{{\includegraphics[width=0.48\linewidth]{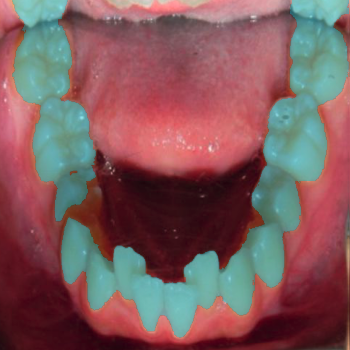} }\label{our-result-bot}}
    \hfill
    \subfloat[\centering Ground-truth mask of lower jaw.]{{\includegraphics[width=0.48\linewidth]{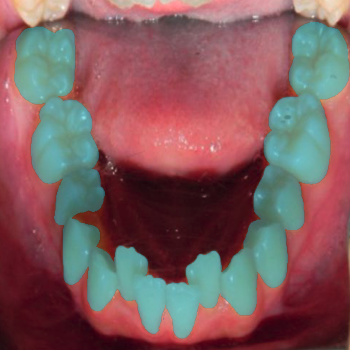} }\label{gt-bot}}%

    \subfloat[\centering Prediction on front view.]{{\includegraphics[width=0.48\linewidth]{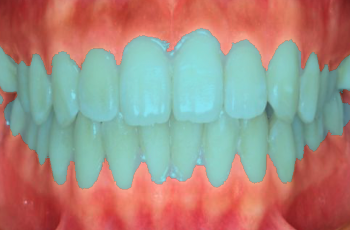} }\label{our-result-front}}%
    \subfloat[\centering Ground-truth mask of front view.]{{\includegraphics[width=0.48\linewidth]{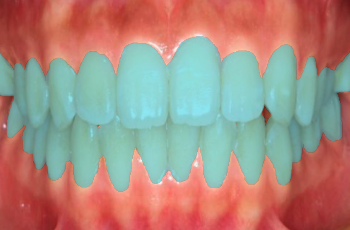} }\label{gt-front}}%

    \subfloat[\centering Prediction on upper jaw.]{{\includegraphics[width=0.48\linewidth]{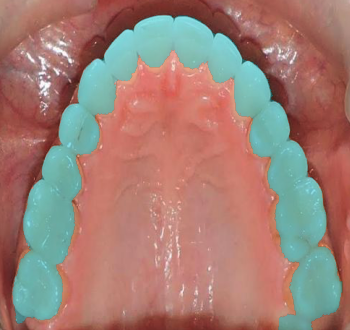} }\label{our-result-top}}
    \hfill
    \subfloat[\centering Ground-truth mask of upper jaw.]{{\includegraphics[width=0.48\linewidth]{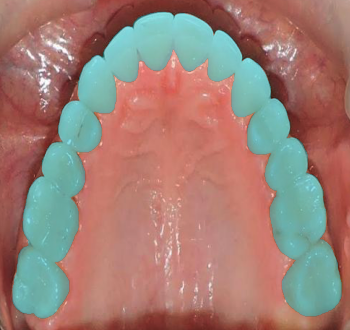} }\label{gt-top}}%
    \caption{Example of results from our proposed solution on compared to the ground truth segmentation masks.\label{fig:our-seg-results}}
\end{figure}

We evaluated the selected methods on the subset of the TriDental dataset which contains ground truth teeth segmentation masks. We report the intersection-over-union metric of the estimated segmentation masks with the ground truth annotations in Table \ref{tab:segmentation_results}. The results show that our method achieved the best results on the images of the lower and frontal view. The Segment Anything model~\cite{SAM} provided with the keypoints from our object detection model achieved the best results on the images from the upper view and tied with our method on the images from the lower view. The remaining methods achieve significantly worse results. 

These findings show that our trained keypoint detection network can efficiently guide our proposed segmentation method to achieve superior results. We also show the results of our proposed segmentation method in Fig.~\ref{fig:our-seg-results}. Our method is successful in providing accurate masks of teeth without the need for costly dataset with mask annotations. 

We have also shown that the keypoints estimated by our keypoint detection network can significantly increase the accuracy of the supervised Segment Anything model. Thus providing a differnt approach which does not require any specialized teeth segmentation dataset to achieve good segmentation results.

\section{Conclusion}

In this research, we have presented a weakly supervised approach for teeth segmentation in dental image analysis. Our method leverages the output heatmaps and intermediate feature maps from a keypoint detection network to guide the segmentation process. By training the keypoint detection network on the TriDental dataset, we have shown that it can effectively discern significant features of teeth. Our multi-scale feature fusion technique enables accurate teeth segmentation without the need for explicit segmentation annotations. The detected keypoints further enhance the quality of the obtained segmentation masks. Experimental results on the TriDental dataset have demonstrated the superiority of our approach in terms of accuracy and robustness compared to existing segmentation methods. Our method offers a cost-effective and efficient solution for teeth segmentation in real-world dental applications, eliminating the need for extensive manual annotation efforts. Future work can focus on expanding the dataset and exploring the generalizability of our method to other dental image datasets, as well as investigating its performance on real-time dental applications.

\section*{Acknowledgment}

This publication is the result of support under the Operational Program Integrated Infrastructure for the project: Advancing University Capacity and Competence in Research, Development and Innovation (ACCORD, ITMS2014+:313021X329), co-financed by the European Regional Development Fund. The work presented in this paper was carried out in the framework of the TERAIS project, a Horizon-Widera-2021 program of the European Union under the Grant agreement number 101079338.

\bibliographystyle{IEEEtran}
\bibliography{literatura} 
\end{document}